\documentclass[sn-basic,iicol,fleqn]{sn-jnl}

\usepackage{graphicx}%
\usepackage{multirow}%
\usepackage{amsmath,amssymb,amsfonts}%
\usepackage{amsthm}%
\usepackage{mathrsfs}%
\usepackage[title]{appendix}%
\usepackage{xcolor}%
\usepackage{textcomp}%
\usepackage{manyfoot}%
\usepackage{booktabs}%
\usepackage{algorithm}%
\usepackage{algorithmicx}%
\usepackage{algpseudocode}%
\usepackage{listings}%
\usepackage{orcidlink}
\usepackage{url}

\theoremstyle{thmstyleone}%
\theoremstyle{thmstyletwo}%

\theoremstyle{thmstylethree}%

\begin{document}

\title[Article Title]{Robust and Flexible Omnidirectional Depth Estimation with Multiple 360-degree Cameras}

\author[1,2]{\fnm{Ming} \sur{Li} \orcidlink{0000-0002-1341-5585}}\email{mingli@smail.nju.edu.cn}
\author[3]{\fnm{Xuejiao} \sur{Hu}}\email{huxuejiao@jit.edu.cn}
\author[1]{\fnm{Xueqian} \sur{Jin}}\email{jcboxq@smail.nju.edu.cn}
\author[1]{\fnm{Jinghao} \sur{Cao}}\email{602022230006@smail.nju.edu.cn}
\author*[1,4]{\fnm{Yang} \sur{Li}}\email{yogo@nju.edu.cn}
\author*[1]{\fnm{Sidan} \sur{Du}}\email{coff128@nju.edu.cn}

\affil*[1]{\orgdiv{School of Electronic Scinence and Engineering}, \orgname{Nanjing University}, \orgaddress{\city{Nanjing}, \postcode{210023}, \country{China}}}
\affil[2]{\orgdiv{School of Artificial Intelligence}, \orgname{anjing University of Information Science and Technology}, \orgaddress{\city{Nanjing}, \postcode{210044}, \country{China}}}
\affil[3]{\orgdiv{School of Computer Engineering}, \orgname{Jinling Institute of Technology}, \orgaddress{\city{Nanjing}, \postcode{211112}, \country{China}}}
\affil[4]{\orgdiv{Suzhou High Technology Research Institute}, \orgname{Nanjing University}, \orgaddress{\city{Suzhou}, \postcode{215123}, \country{China}}}








\abstract{Omnidirectional depth estimation has received much attention from researchers in 3D perception and measurement in recent years. However, challenges arise due to camera soiling and variations in camera layouts, affecting the robustness and flexibility of the algorithm. In this paper, we use the geometric constraints and redundant information of multiple 360$^\circ$ cameras to achieve robust and flexible multi-view omnidirectional depth estimation. We implement two algorithms, in which the two-stage algorithm obtains initial depth maps by pairwise stereo matching of multiple cameras and fuses the multiple depth maps to achieve the final depth estimation; the one-stage algorithm adopts spherical sweeping based on hypothetical depths to construct a uniform spherical matching cost of the multi-camera images and obtain the depth. Additionally, a generalized epipolar equirectangular projection is introduced to simplify the spherical epipolar constraints. To overcome panorama distortion, a spherical feature extractor is implemented. Furthermore, a synthetic 360$^\circ$ dataset on outdoor road scenes is presented to train and evaluate 360$^\circ$ depth estimation algorithms. Our dataset takes soiled camera lenses and glare into consideration, which is more consistent with the real-world environment. Experiments show that our two algorithms achieve state-of-the-art performance, accurately predicting depth maps even when provided with soiled panorama inputs. The flexibility of the algorithms is experimentally validated in terms of camera layouts and numbers.}




\maketitle


\section{Introduction}\label{sec1}
Vision-based depth estimation is an essential method for 3D environmental perception. Recently, omnidirectional depth estimation has attracted attention in numerous applications including autonomous driving and robot navigation, owing to its efficiency of the 360$^\circ$ environment. Various algorithms have been proposed to estimate omnidirectional depth maps, including monocular \cite{Wang2020bifuse,Jiang2021unifuse,Li2022omnifusion}, binocular \cite{Li2021csdnet,360SD-Net} and multi-view approaches \cite{SweepNet,Omnimvs,Omnimvs_J,Su2023omni}.

\begin{figure}[t]%
  \centering
  \includegraphics[width=0.5\textwidth]{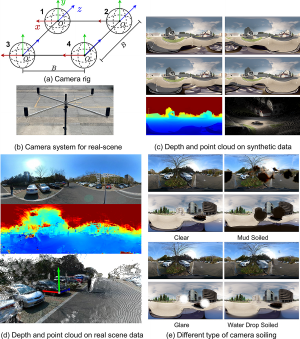}
  \caption{Overview of the proposed robust and flexible multi-view omnidirectional depth estimation framework. (a) and (b) show the multiple 360$^\circ$ camera rig. (c) and (d) show the results of predicted depth map and reconstructed point cloud on synthetic and real-world data. (e) illstrates the different type of camera soiling in practice. For each sample in (e), the upper and the lower show the soiled panoramas in real-world and synthetic dataset, respectively}\label{fig_overview}
\end{figure}

The complex geometric constraints and image distortions of spherical images pose challenges for omnidirectional depth estimation. In addition, the camera may be soiled resulting in image degradation in practical applications. As shown in Fig. \ref{fig_overview}(e), the images can be soiled by mud spots, water drops or dazzled by intense light. Besides, the camera layouts may vary to accommodate different types of vehicles or robots in real-world tasks. Consequently, the development of an omnidirectional depth estimation algorithm that exhibits robustness against camera soiling and flexibility in adapting to diverse camera configurations becomes imperative and indispensable.

However, most of existing methods either extract spherical features with conventional planar convolution\cite{Wang2020bifuse,Jiang2021unifuse,360SD-Net} or do not simplify the spherical epipolar constraint\cite{Li2021csdnet}. Apart from this, monocular omnidirectional depth estimation methods susceptible to overfitting the scenes of the training data and are unable to mitigate the impact of camera soiling. Binocular methods also encounter challenges in obtaining reliable depth maps when 360$^\circ$ cameras installed on vehicles are soiled. Won et al. proposed multi-view methods SweepNet\cite{SweepNet} and OmniMVS\cite{Omnimvs,Omnimvs_J} to estimate 360$^\circ$ depth maps from four fisheye cameras. However, these methods also use planar convolution to extract spherical features, and have not used multiple cameras infomation to improve the robustness.

In this paper, we propose the Generalized Epipolar Equirectangular (GEER) projection, which simplifies the geometric constraints of binocular spherical images, enabling the definition of disparity and cost construction for spherical stereo matching. Moreover, OmniMVS\cite{Omnimvs} introduces the spherical sweeping method to establish multi-view spherical geometric constraints. By applying these two types of geometric constraint models, we propose two multi-view omnidirectional depth estimation(MODE) algorithms.

The first method, termed Pairwise Stereo MODE (PSMODE), employs a two-stage approach for multi-view omnidirectional depth estimation. In the first stage, we choose several camera pairs from different views for omnidirectional stereo matching and obtain disparity maps. In the second stage, we convert these disparity maps to aligned depth maps and fuse them to estimate the final depth. Inspired by MVSNet\cite{Mvsnet} and OmniMVS\cite{Omnimvs}, we leverage Spherical Sweeping and construct a unified cost volume for multi-view panoramas to implement the one-stage SSMODE method. SSMODE generates the cost volume by sweeping the hypothetical spheres at different depths and aggregates the cost to obtain 360$^\circ$ depth maps. Additionally, we introduce a spherical feature extraction module to mitigate the distortion present in panoramas.\footnotemark[1]
\footnotetext[1]{We use the terms omnidirectional, 360$^\circ$, spherical and panorama interchangeably in
  this document.}
Moreover, a large-scale synthetic outdoor omnidirectional dataset, Deep360, is proposed in this work. To evaluate the performance of different 360$^\circ$ depth estimation methods when camera lenses are soiled by mud spots, water drops or dazzled by glare, we also provide a soiled version of the dataset.

Experimental results demonstrate that both two methods generate reliable depth maps in various scenes and achieve state-of-the-art (SOTA) performance on different datasets, especially the one with soiled panoramas. This validates the robustness of our proposed frameworks. In addition, we evaluate the two methods on datasets  featuring diverse camera settings and varying numbers of cameras to demonstrate the flexibility of our frameworks, which can be extended to arbitrary 360$^\circ$ multi-camera configurations. We also present a comprehensive comparison of two types of spherical geometry constraint models and two depth estimation algorithms.

In summary, the main contributions of this work are as follows:

\begin{itemize}
  \item We leverage the geometric constraints and redundant information of multiple 360$^\circ$ cameras to achieve robust and flexible multi-view omnidirectional depth estimation. To this end, we introduce two methods that adopt pairwise stereo matching and spherical sweeping, respectively. Experiments show that both two methods achieve state-of-the-art performance. We demonstrate that the proposed methods are robust against camera soiling and flexible with different camera layouts by extensive experiments. A comprehensive comparison of two types of spherical geometry constraint models and algorithms is also presented in this paper.
  \item We introduce the spherical convolution to mitigate panorama distortions in 360$^\circ$ stereo matching. We propose the Generalized Epipolar Equirectangular projection for 360 camera stereo pairs at arbitrary relative positions to leverage the epipolar constraint.
  \item We present a large-scale synthetic outdoor dataset, Deep360, that contains both high-quality and soiled panorama images.
\end{itemize}

Compared to our conference version\cite{Li2022mode}, this extended work encompasses following advancements. Firstly, we expand the applicability of the Cassini projection, to the Generalized Epipolar Equirectangular projection, which accommodates camera pairs at arbitrary relative positions. We provide a thorough analysis and comparison of the spherical geometry constraint models. We introduce the one-stage Spherical-Sweeping MODE and extensively compare its performance with the two-stage Pairwise Stereo matching methods through a wealth of experiments. Furthermore, we demonstrate the flexibility of the proposed methods with varying layouts and numbers of input cameras. Lastly, we present a comprehensive comparative analysis, encompassing the latest state-of-the-art methods, and provide insights for the future advancement of the field.

\section{Related Work}\label{sec2}
\subsection{Stereo Matching and Multi-view Stereo Methods}\label{sec2_1}
Conventional stereo matching methods estimate disparity map based on the stereo epipolar constraint and image features matching. Some methods aggregate global features to achieve high accuracy, such as SGM\cite{Hirschmuller2005sgm} and its variants\cite{Hu2012wdpsm,Michael2013opsgm,Li2019ddrsgm}, and graph-cut based methods\cite{Boykov2001gc,Boykov2004gc2}. Deep learning methods report much improved performance in stereo matching. Zbontar and Lecun propose MCCNN\cite{zbontar2016mccnn} that implements the feature extraction with CNNs and computes disparity via conventional cost aggregation. Many methods\cite{GC-Net,PSMNet,Zhang2019GANet,Shen2021CFNet,Xu2022acvnet,Chong2023SA-Net} construct 3D cost volume with image features and optimize the 3D-CNN based cost aggeration modules to estimate disparity maps. Some approaches\cite{Mayer2016dispnet,Pang2017crl,Zheng2024stereo_MIS} compute the 2D left-right feature correlation volume. AANet\cite{Xu2020aanet} adopts an adaptive aggregation algorithm and replaces the costly 3D-CNNs for an efficient architecture. DMCA-Net\cite{Zeng2024DMCA-Net} utilizes differentiable Markov Random Field for cost aggeration to guide stereo matching. RAFT-Stereo\cite{lipson2021raft} adopts multi-level Gated Recurrent Unit (GRU) to estimate disparity maps recurrently. CREStereo\cite{Li2022crestereo} designs a hierarchical network to update disparities iteratively and proposes an adaptive group correlation layer to match points via the local feature.

Multi-view Stereo (MVS) has important applications in 3D reconstruction and has developed rapidly in recent years. Yao et al. \cite{Mvsnet} proposed the end-to-end MVSNet that builds cost volume by warping feature maps of different views into front-parallel planes of the reference camera to obtain depth maps. P-MVSNet\cite{Luo2019pmvsnet} proposes a patch-wise aggregation to build confidence volume and a hybrid network of isotropic and anisotropic 3D-CNNs to exploit context information. Point-MVSNet\cite{Chen2019pointmvs} adopts the feature augmented point cloud to refine the depth map iteratively. Cascade-MVS\cite{Gu2020cascademvs} and CVP-MVS\cite{Yang2021cvpmvs} improve the performance with multi-scale coarse-to-fine architectures. UGNet\cite{Su2022UGNet} also adopts a coarse-to-fine architecture and leverages uncertainty to improve the depth accuracy. DS-Depth\cite{Miao2024DS-Depth} builds the fusion cost volume from multi-frame images to estimate accurate depth maps. PVA-MVSNet\cite{Yi2020pvamvsnet} proposes self-adaptive view aggregation to generate cost volume instead of the widely-used mean square variance. PVSNet\cite{Xu2022invmvs} and Vis-MVSNet\cite{Zhang2023vismvs} take the visibility of each view into consideration to supress the mis-matching. Many approaches use the iterative optimization modules to replace the 3DCNNs. R-MVSNet\cite{Yao2019rmvsnet} and CER-MVS\cite{Ma2022cermvs} adopt the GRU module and D2HC-RMVSNet\cite{Yan2020lstmmvs} leverages the LSTM module for the cost aggregation. Chen et al.\cite{Chen2025STViT+} propose a spatial-temporal transformer and leverage self-supervised scheme for multi-view multi-frame depth estimation.

These stereo matching methods are designed for perspective cameras with normal field-of-view (FoV) and do not consider the property of panoramas.

\subsection{Omnidirectional Depth Estimation}\label{sec2_2}
Omnidirectional depth estimation has attracted the attention of researchers because of the efficient perception for 360$^\circ$ surrounding environment. Shih et al. propose a stereo vision system based on two omnidirectional cameras\cite{Shih2013omnisystem,Shih2013omnisystem-2}. Recently, many learning-based algorithms have been proposed. Zioulis et al. propose two monocular networks using supervised learning\cite{OmniDepth}, and adopt the extra coordinate feature in CoordNet\cite{3D60} for learning context in the equirectangular projection (ERP) domain. Some algorithms solve the distortion problem of panorama with projection transformation. Wang et al\cite{PanoSUNCG} proposed a self-supervised framework to estimate omnidirectional depth and camera poses from 360 videos. They further propose BiFuse\cite{Wang2020bifuse} for monocular depth estimation which combines the ERP and CubeMap projection to overcome the distortion of panoramas. Jiang et al. also develop the fusion scheme and propose UniFuse\cite{Jiang2021unifuse} which achieves better performance via a more efficient fusion module. BiFuse++\cite{Wang2023bifuse2} integrates the bi-projection fusion architecture into self-supervised monocular 360$^\circ$ depth estimation and improves the fusion module. SegFuse\cite{Feng2022segfuse} also proposes a two-branch network to fuse the features of ERP and CubeMap projection images and predicts the omnidirectional depth and semantic segmentation maps. OmniFusion\cite{Li2022omnifusion} transforms the panorama into less-distorted perspective patches and merge the patch-wise depth predictions for the omnidirectional depth map. Cheng et al.\cite{Cheng2020odecnn} regard omnidirectional depth estimation as an extension of the partial depth map. Some methods estimation omnidirectional depth maps from binocular panoramic images. Wang et al.\cite{360SD-Net} propose the 360SD-Net which follows the stereo matching pipeline to estimate omnidirectional depth in the ERP domain for up-down stereo pairs. CSDNet\cite{Li2021csdnet} focuses on the left-right stereo and uses Mesh CNNs to solve the spherical distortions and proposes a cascade framework to estimate accurate depth maps. However, these methods either extract spherical features with planar convolution or do not simplify the spherical epipolar constraint.

There are also some methods for obtaining omnidirectional depth maps based on multi-view fisheye cameras. Won et al. propose SweepNet\cite{SweepNet} which builds cost volume via spherical sweeping and estimates spherical depth by cost aggregation. They further improve the algorithm and propose the end-to-end OmniMVS\cite{Omnimvs,Omnimvs_J} architecture to achieve better performance. Meuleman et al. \cite{realtimeOmni} propose an adaptive spherical matching method and an efficient cost aggregation method to achieve real-time omnidirectional MVS. Yang et al. \cite{YangKL22Dense360depth} introduce a translation scaling scheme to extend the spherical camera model to multiview for dense 360$^\circ$ depth. OmniVidar\cite{OmniVidar} adopts the triple sphere camera model and rectifies the multiple fisheye images into stereo pairs of four directions to obtain depth maps. Su et al.\cite{Su2023omni} leverage a cascade architecture for cost regularization to achieve high accuracy for omnidirectional detph extimation from four fisheye cameras. However, these methods also use planar convolution to extract spherical features and the blind areas of fisheye cameras introduce discontinuity in the spherical cost volume.

\subsection{Omnidirectional Depth Datasets}\label{sec2_3}
Large-scale datasets with high variety are essential for training and evaluating learning-based algorithms. Recently released omnidirectional depth datasets can be divided into two categories according to the input images, one with the panoramas, and the other with the fisheye images. These datasets are mainly collected from publicly available real-world and synthetic 3D datasets by repurposing them to omnidirectional by rendering. For datasets with panoramas, Wang et al.\cite{PanoSUNCG} collect an indoor monocular 360$^\circ$ video dataset named PanoSUNCG from\cite{SUNCG}. De La Garanderie et al.\cite{MonoCarla} provide an outdoor monocular 360$^\circ$ benchmark with 200 images generated from the CARLA autonomous driving simulator\cite{CARLA}. MP3D and SF3D\cite{360SD-Net} are indoor binocular 360$^\circ$ datasets collected from\cite{Matterport3D,Stanford3D}. 3D60 by Zioulis et al.\cite{3D60} is an indoor trinocular (central, right, up) 360$^\circ$ dataset collected from\cite{Matterport3D,Stanford3D,SUNCG,SceneNet}. For datasets with fisheye images, Won et al.\cite{SweepNet,Omnimvs,Omnimvs_J} present three datasets: Urban, OmniHouse and OmniThings. All three datasets are virtually collected in Blender with four fisheye cameras. The fisheye images need complementary information to estimate an omnidirectional depth map, which means discontinuity and requirements for camera directions. In contrast, the panoramas record all 360$^\circ$ information continuously without blind areas. However, as summarized above, the datasets with stereo panoramas consist of indoor scenes only. A detailed summary of multi-view omnidirectional depth datasets can be found in Table \ref{tab_dataset}.

\section{Spherical Geometry Constraint Model}\label{sec3}
\begin{figure}[t]%
  \centering
  \includegraphics[width=0.4\textwidth]{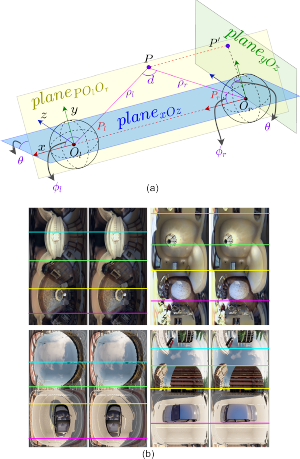}
  \caption{(a) The coordinate definition and geometry of the proposed generalized epipolar equirectangular projection. (b) The samples of omnidirectional stereo pairs at different relative poses on GEER projection. The spherical epipolar constraint is simplified to horizontal lines on GEER projection}\label{fig_projection}
\end{figure}
To achieve the robust and accurate depth estimation, we establish the geometry constrint of multiple 360$^\circ$ cameras. In this paper, we introduce two spherical geometry constraint models to leverage the multi-view information. In Section \ref{sec3_1}, we introduce the generalized epipolar equirectangular projection, which simplifies the epipolar constraint for binocular panoramas and enables the stereo matching methods on spherical images. In Section \ref{sec3_2}, we present the pipeline of spherical sweeping that builds the cost volume of multi-view panoramas based on the hypothetical sphericals.

\subsection{Generalized Epipolar Equirectangular Projection}\label{sec3_1}
Equirectangular projection (ERP) is widely used to represent spherical images. ERP linearly represents the latitude and longitude in spherical coordinates as pixel coordinates and projects the panorama into the planar image. 360SD-Net\cite{360SD-Net} estimates the disparity map of up-down omnidirectional stereo pairs in ERP domain. Li et al.\cite{Li2008fisheye} proposed latitude-longtitude projection to build epipolar constraint for left-right spherical stereo. \cite{Li2022mode} adopts Cassini projection\footnotemark[2] for left-right omnidirectional stereo matching. These projection methods also linearly represent the angle coordinates on the sphere as pixel coordinates on the image, using a rotated coordinate definition with ERP.
\footnotetext[2]{\url{https://en.wikipedia.org/wiki/Cassini_projection}}

In this paper, we propose Generalized epipolar equirectangular (GEER) projection to achieve the epipolar constraint for binocular panoramic cameras at arbitrary relative positions in space. As shown in Fig. \ref{fig_projection}(a), $O_l$ and $O_r$ are the optic centers of two omnidirectional cameras. We establish a 3D Cartesian coordinate system, where the direction of the x-axis is $O_rO_l$. $P$ is an object point in 3D space and imaged at points $P_l$ and $P_r$ on the left and right imaging spheres respectively. $P'$ is the projection of $P$ on the plane $yOz$. We define the spherical coordinate system $(\rho, \phi,\theta)$ as follows: $\rho$ is the distance between the object point $P$ and the optic center $O$, $\phi$ is the angle between $PO$ and $x$ axis ($\angle POx$) and denotes the elevation angle, $\theta$ is the angle between $P'O$ and $z$ axis on the plane $yOz$ ($\angle P'Oz$) and denotes the azimuth angle. Thus, the transformation between Cartesian coordinates and the spherical coordinates is:
\begin{align}
  \left\{
  \begin{aligned}
    x & =  \rho\cos(\phi)                   \\
    y & =  \rho\sin(\phi)sin(\theta),\qquad \\
    z & =  \rho\sin(\phi)cos(\theta)
  \end{aligned}
  \left\{
  \begin{aligned}
    \rho   & = \sqrt{(x^{2}+y^{2}+z^{2})} \\
    \phi   & =arccos({\dfrac{x}{\rho}})   \\
    \theta & =arctan({\dfrac{y}{z}})
  \end{aligned}
  \right.
  \right.\label{eq1}
\end{align}
where $\phi \in \left[0, \pi\right], \theta \in \left[-\pi, \pi\right]$. The points on the sphere are projected to the images with the mapping function:
\begin{align}
  \left\{
  \begin{aligned}
    u & =  \phi \cdot \dfrac{W}{\pi}          \\
    v & =  (\theta +\pi)\cdot \dfrac{H}{2\pi}
  \end{aligned}
  \right.
  \label{eq2}
\end{align}
where $(u,v)$ denotes the image pixel coordinates in GEER projection and $H,W$ denote the height and width of the image. Because $\theta (\angle P'Oz)$ also denotes the angle between the plane $P O_l O_r$ and the plane $xOz$, the imaging points $P_l$ and $P_r$ have the same $\theta$ value in the spherical coordinate. Thus, $P_l$ and $P_r$ have the same vertical coordinate $u$ on GEER projection images. In other words, the epipolar lines are projected to horizontal lines in GEER domain. As shown in Fig. \ref{fig_projection}(b), although the image structures of projection maps are different for different camera rigs, the matching points in stereo images lie on the same horizontal lines. Therefore, with GEER projection we can transform the two panoramas at arbitrary relative position into left-right stereo pairs that follows the epipolar constraint.
Since the matching points have the same $\theta$, the angular disparity $d$ is defined as the difference of $\phi$ : $d=\phi_l - \phi_r$. The depth of $P$ to the left camera is computed as:

\begin{equation}
  \rho_l=B\cdot \dfrac{sin(\phi_r)}{sin(d)}=B\cdot \left[\dfrac{sin(\phi_l)}{tan(d)}-cos(\phi_l)\right].
  \label{eq3}
\end{equation}

\subsection{Spherical Sweeping}\label{sec3_2}
\begin{figure}[t]%
  \centering
  \includegraphics[width=0.4\textwidth]{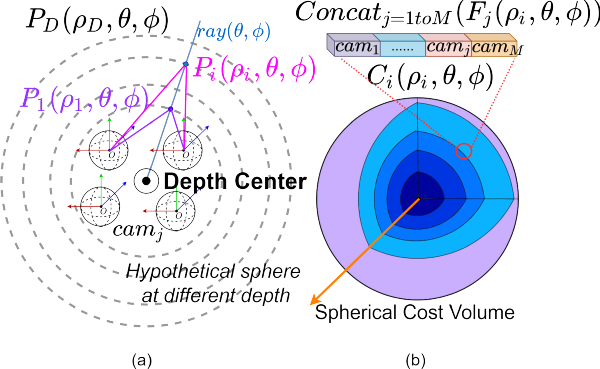}
  \caption{(a) The process of spherical sweeping. (b) The construction of the spherical cost volume. The points at different hypotheses depth can be projected to the cameras coordinates to obtain the features. Then the features of the same point from different cameras are concatenated to represent the matching cost}\label{fig_sphere_sweep}
\end{figure}

We define the disparity of binocular 360$^\circ$ cameras with the GEER projection. Thus, the exsisting stereo matching approaches can be applied to spherical images. However, disparity can only represent the geometry of two cameras. To leverage information of multiple cameras, we need to adopt the stereo matching for different camera pairs.

Inspired by the MVSNet\cite{Mvsnet} and OmniMVS\cite{Omnimvs}, we utilize the spherical sweeping method to build the unified cost volume with multi-view panoramas. As illustrated in Fig. \ref{fig_sphere_sweep}, we construct a series of hypothetical spheres at different depths. According to Equation(\ref{eq2}), each pixel in the target depth map can be envisioned as representing a ray of light in space ($ray(\theta, \phi)$), and associating it with different depths corresponds to different potential object points along that ray. For each point $P_i(\rho_i,\theta,\phi)$, we can find the corresponding image coordinates of each camera:
\begin{equation}
  (u_{ij}, v_{ij})_{\theta,\phi} = K_jT_jP_i(\rho_i,\theta,\phi)
  \label{eq4}
\end{equation}
where $\rho_i$ denotes the hypothetical depth at index $i$, $K_j$ and $T_j$ denote the intrinsic and extrinsic matrix of the camera with the index $j$.
To build the matching cost of point $P_i(\rho_i,\theta,\phi)$, we concatenate the features from different views:
\begin{equation}
  C_i(\rho_i,\theta,\phi) = Concat_{j=1}^M(F_j(u_{ij}, v_{ij})_{\theta,\phi})
  \label{eq5}
\end{equation}

For the point at the hypothetical depth that close to the real depth value, the features from different cameras are more consistency compared to other hypothetical depths. Thus, the geometry constraint of multiple cameras is established based on the spherical sweeping.

In this paper, we introduce two omnidirectional depth estimation mthods that establish geometry constraint based on GEER and Spherical Sweeping method, respectively. We introduce the two algorithms separately in Section \ref{sec4}. Subsequently, we conduct comprehensive experiments to validate and compare the performance of these two methods.

\section{Method}\label{sec4}
We leverage the redundant information and geometry constraint of multiple 360$^\circ$ cameras, and introduce two frameworks to obtain omnidirectional depth maps. We first adopt the GEER projection to apply the epipolar constraint for spherical stereo and propose Pairwise Stereo Multi-view Omnidirectional Depth Estimation (PSMODE), a novel two-stage approach consisting of pairwise stereo matching and depth map fusion. In the first stage, we select several camera pairs from different views for omnidirectional stereo matching and obtain disparity maps. In the second stage, we convert disparity maps to aligned depth maps and fuse them to estimate the final depth map. We further implement the one-stage Spherical Sweeping Multi-view Omnidirectional Depth Estimation (SSMODE) that builds the unified cost volume with spherical sweeping method. SSMODE first extracts features for each panorama, then constructs 360$^\circ$ cost volume through hypothetical spheres of different depths. The costs are aggregated to estimate the depth map.

\subsection{Spherical Feature Extraction Module}\label{sec4_1}

\begin{figure}[t]%
  \centering
  \includegraphics[width=0.45\textwidth]{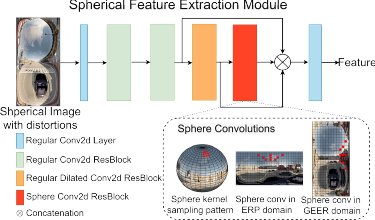}
  \caption{The structure of proposed spherical feature extraction module. We use four stages of residual blocks to build the module and fuse the features from different stages. The sphere convolution is adopted in the last stage to obtain high-level semantic and context features}\label{fig_feature}
\end{figure}

Extracting context features from distorted spherical images is challenging for regular CNN modules. In this paper, we implement a Spherical Feature Extraction Module based on spherical convolutions to mitigate the distortion of panoramas. As shown in Fig. \ref{fig_feature}, we implement the sphere convolution based on\cite{Coors2018spherenet} and accelerate it with CUDA. The sphere convolution changes the sampling pattern to convolve through the neighborhood pixels on the sphere instead of the panorama.

The proposed spherical feature extraction module contains four stages of residual blocks\cite{He2015resnet}. Dilated convolutions are employed in the third stage of residual blocks to facilitate the incorporation of large receptive fields. Spherical convolutions are utilized in the final stage to extract high-level semantic and context features. Our implementation of spherical convolutions can be applied to different spherical map projections such as ERP and proposed GEER projection. The spherical feature extraction module is employed in both two-stage (PSMODE) and one-stage (SSMODE) omnidirectional depth estimation networks.

\subsection{Pairwise Stereo Matching and Depth Fusion (PSMODE)}\label{sec4_2}
We propose a two-stage approach named Pairwise stereo Multi-view Omnidirectional Depth Estimation (PSMODE), which fuses the depth maps estimated via pairwise stereo matching.
\begin{figure*}[t]%
  \centering
  \includegraphics[width=0.8\textwidth]{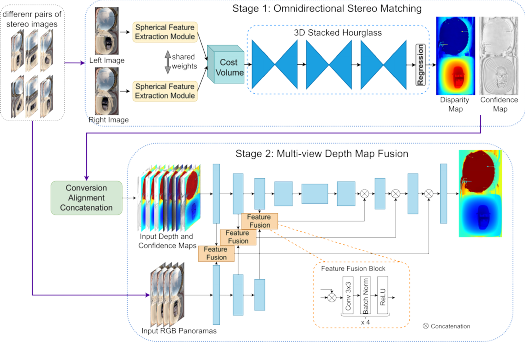}
  \caption{The architecture of proposed PSMODE, which contains two stage to estimate the omnidirectional depth map. In the first stage, we propose an omnidirectional stereo matching network to obtain depth maps and confidence maps of different stereo pairs. In the second stage, we fuse the multi-view depth maps to estimate the final depth maps}\label{fig_psmode}
\end{figure*}

\subsubsection{Pairwise Stereo Matching}\label{sec4_2_1}
Fig. \ref{fig_psmode} illustrates the process involved in PSMODE. Initially, the multi-view panoramas are organized into multiple stereo image pairs, which are subsequently transformed into GEER projections for pairwise stereo matching. To address distortions, the left and right images are passed through the Spherical Feature Extraction Module to generate feature maps. These feature maps are shifted and concatenated to construct the cost volume. A 3D stacked hourglass network is employed to aggregate the cost volume and estimate the disparity map. The network is optimized with the smoothL1 loss function during training.

Moreover, many stereo matching algorithms take a random crop of images as the network input. However, different crop areas on spherical projection images have different distributions in the high-level feature space due to the image distortions. Thus, we use the full omnidirectional images without cropping as the input of the proposed network to achieve better performance.

\subsubsection{Omnidirectional Depth Fusion}\label{sec4_2_2}
In the second stage of PSMODE, the disparity maps are converted to aligned depth maps and fused to estimate the final depth map. To reduce the effect of predicted disparity errors, we add confidence maps into the second stage of PSMODE to provide extra information for the depth map fusion. Poggi et al.\cite{Poggi2021conf} reviews developments in the field of confidence estimation for stereo matching and evaluates existing confidence measures. Considering that the stereo matching network computes each disparity value through a probability weighted sum over all disparity hypotheses, the probability distribution along the hypotheses thus reflects the quality of disparity estimation. We compute the confidence for each inferred disparity value by taking a probability sum over the three nearest disparity hypotheses, which corresponds to the probability that the inferred disparity meets the 1-pixel error requirement.

We align the depth maps and confidence maps to the same viewpoint based on the extrinsic matrix and visibility. As shown in Fig. \ref{fig_psmode}, the depth fusion network generally follows Unet\cite{Unet}, containing two encoders and one decoder. One encoder takes concatenation of the aligned depth maps and confidence maps as input to effectively aggregate the depth feature. and the other takes RGB panoramas as input to extract context and boundary features. Subsequently, these two types of features are fused through a multi-scale feature fusion block to generate the more comprehensive and informative feature maps. Finally, the decoder utilizes fused feature maps to perform regression and predict the final depth map.

We adopt the training loss developed from Scale-Invariant Error (SILog)\cite{Silog} as:
\begin{align}
  Loss(\hat{y}, y^\star) & = \frac{1}{n}\sum_i d_i^2-\frac{\lambda}{n^2} {\left({\sum_i d_i}\right)}^2 \\
  d_i                    & = \log \hat{y}_i - \log y_i^\star \label{eq6}
\end{align}
where $\hat{y}$ denotes the predicted depth map and $y^\star$ denotes the ground truth and $\lambda \in \left[0, 1\right]$. We follow \cite{Silog} to set $\lambda = 0.5$ in the experiments, which averages the scale-invariant depth error and absolute-scale error.

\subsection{Spherical Sweeping Multi-view Omnidirectional Depth Estimation(SSMODE)}\label{sec4_3}
\begin{figure*}[t]%
  \centering
  \includegraphics[width=0.8\textwidth]{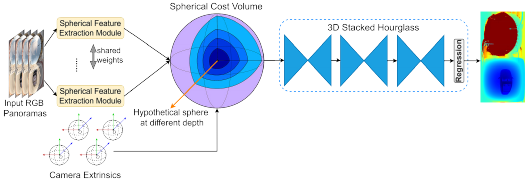}
  \caption{The architecture of proposed SSMODE. We build the unified spherical cost volume with hypothetical sphere at different depth to predict the omnidirectional depth maps}\label{fig_ssmode}
\end{figure*}

Inspired by Multi-view Stereo(MVS) and OmniMVS \cite{Omnimvs}, we leverage spherical sweeping to build the unified spherical cost volume for multi-view panoramas and propose the Spherical Sweeping MODE(SSMODE).

As shown in Fig. \ref{fig_ssmode}, the proposed SSMODE first extracts features of input panoramas with the Spherical Feature Extraction Module. By employing the GEER projection, we establish the angular coordinate system $(\theta,\phi)$ and define a collection of hypothetical spheres at different depths $\rho_i$. The features for each point $(\rho_i,\theta,\phi)$ across different views are obtained through the camera extrinsics, and these features are concatenated to construct the spherical cost volume. As illustrated in Fig. \ref{fig_sphere_sweep} and Equation(\ref{eq5}), the unified spherical cost volume contains the matching cost of each pixel at each hypothetical depth. Similar to stereo matching, the spherical cost volume is also represented as a 5D tensor with a shape of $(B \times C \times D \times H \times W)$, where $(H \times W)$ denotes the angular coordinate of the sphere and $D$ represetns the number of hypothetical spheres. Subsequently, the 3D stacked hourglass module is employed to aggregate the multi-view spherical matching cost. Based on the regressed comprehensive matching cost, the weights of different depth are calculated, and the final depth is obtained by weighted summation:
\begin{align}
  w_i(\rho_i,\theta,\phi)=\frac{e^{C'_i(\rho_i,\theta,\phi)}}{\sum_{i=1}^{D}e^{C'_i(\rho_i,\theta,\phi)}} \\
  depth(\theta,\phi)=\sum_{i=1}^{D}w_i\rho_i\label{eq7}
\end{align}

To overcome distortions, we utilize the GEER projection to represent the input panoramas and employ the Spherical Feature Extraction Module. During training, SSMODE is optimized using the multi-stage smoothL1 loss function, as presented in PSMNet\cite{PSMNet}.

\section{Dataset}\label{sec5}
\begin{table*}[]
  \center
  \caption{Overview of the proposed datasets and other published datasets}
  \label{tab_dataset}
  \resizebox{\textwidth}{!}{%
    \begin{tabular}{cccccccc}
      \toprule
      \multicolumn{2}{c}{Datasets}                 & Scene          & Input          & Views    & Training & Testing & Validation        \\ \midrule
      \multirow{3}{*}{Won et.al\cite{Omnimvs}}     & Urban          & Outdoor        & fisheye  & 4        & 700     & 300        & N/A  \\
                                                   & OmniHouse      & Indoor         & fisheye  & 4        & 2048    & 512        & N/A  \\
                                                   & OmniThings     & Random objects & fisheye  & 4        & 9216    & 1024       & N/A  \\ \midrule
      \multirow{2}{*}{Wang et. al\cite{360SD-Net}} & SF3D           & Indoor         & panorama & 2        & 800     & 203        & 200  \\
                                                   & MP3D           & Indoor         & panorama & 2        & 1602    & 341        & 431  \\ \midrule
      Zioulis et. al\cite{3D60}                    & 3D60           & Indoor         & panorama & 3        & 7858    & 2190       & 1103 \\ \midrule
      \multirow{2}{*}{Ours}                        & Deep360        & Outdoor        & panorama & 4        & 2100    & 600        & 300  \\
                                                   & Deep360-soiled & Outdoor        & panorama & 4        & 2100    & 600        & 300  \\ \bottomrule
    \end{tabular}%
  }
\end{table*}

\begin{figure}[t]%
  \centering
  \includegraphics[width=0.5\textwidth]{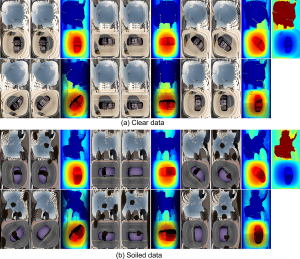}
  \caption{(a) and (b) show the sample of clear data and soiled data of proposed Deep360. Each frame contains 6 different pairs of stereo panoramas in GEER domain, 6 corresponding disparity maps and one depth map}\label{fig_dataset}
\end{figure}

As summarized in \ref{sec2_3}, although many datasets have been proposed for omnidirectional depth estimation, no 360$^\circ$ stereo dataset for outdoor road scenes is available due to the difficulty of acquiring 360$^\circ$ outdoor 3D datasets in the real world. Therefore, we create a public available 360$^\circ$ multi-view dataset Deep360 based on the CARLA autonomous driving simulator. Fig. \ref{fig_dataset} shows some examples of the dataset. We set four 360°cameras and arrange the cameras on a horizontal plane to form a square with side length as one meter, as shown in Fig. \ref{fig_overview}(a). The cameras are numbered from 1 to 4. Any two of the cameras can form a stereo pair, so there are 6 $(C_4^2)$ pairs in total. Each frame consists of six pairs of rectified panoramas, which cover all the pairwise combinations of four 360$^\circ$ cameras, six corresponding disparity maps and one ground truth depth map. All these images and maps have a resolution of $1024 \times 512$.
To acquire realistic 360$^\circ$ outdoor road scenes with high variety, we make the car with 360$^\circ$ cameras in CARLA drive automatically \cite{CARLA} in six different towns and spawn many other random pedestrian and vehicles.

We also provide a soiled version of the Deep360 dataset, which can be used to train and evaluate 360$^\circ$ depth estimation algorithms under harsh circumstances in autonomous driving. Deep360-Soiled contains panoramas soiled or affected by three common outdoor factors: mud spots, water drops and glare, as illustrated in Fig. \ref{fig_overview}(e). An overview of the proposed dataset and other published 360$^\circ$ datasets is listed in Table \ref{tab_dataset}.

\section{Experiment Results}\label{sec6}

\begin{table*}[]
  \center
  \setlength{\tabcolsep}{1mm}
  \caption{Quantitative results of stereo matching on the proposed Deep360 dataset. The metrics refer to disparity errors}
  \label{tab_res_disp}
  \resizebox{0.6\textwidth}{!}{%
    \begin{tabular}{ccccccc}
      \toprule
      \multirow{2}{*}{Methods}        & \multicolumn{6}{c}{Metrics}                                                                                                           \\ \cmidrule{2-7}
                                      & MAE$\downarrow$             & RMSE$\downarrow$ & Px1(\%)$\downarrow$ & Px3(\%)$\downarrow$ & Px5(\%)$\downarrow$ & D1(\%)$\downarrow$ \\ \midrule
      PSMNet\cite{PSMNet}             & 0.3501                      & 1.8244           & 4.3798              & 1.3559              & 0.8398              & 1.2973             \\
      AANet\cite{Xu2020aanet}         & 0.5057                      & 2.2232           & 7.7282              & 2.0914              & 1.1887              & 1.7929             \\
      360SD-Net\cite{360SD-Net}       & 0.4235                      & 1.8320           & 6.6124              & 1.9080              & 1.0885              & 1.7753             \\
      CREStereo\cite{Li2022crestereo} & 0.2779                      & 1.5529           & 3.9118              & 1.4471              & 0.8753              & 1.3088             \\
      Ours                            & \textbf{0.2073}             & \textbf{1.2347}  & \textbf{2.6010}     & \textbf{0.8767}     & \textbf{0.5260}     & \textbf{0.8652}    \\ \bottomrule
    \end{tabular}%
  }
\end{table*}

\begin{figure}[t]%
  \centering
  \includegraphics[width=0.45\textwidth]{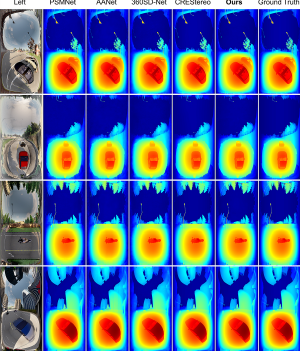}
  \caption{Comparison of the qualitative results of the proposed omnidirectional stereo matching method with other representative binocular stereo matching methods. We show the results in GEER projection since the spherical disparity is defined in GEER domain}\label{fig_res_disp}
\end{figure}

\begin{figure}[t]%
  \centering
  \includegraphics[width=0.5\textwidth]{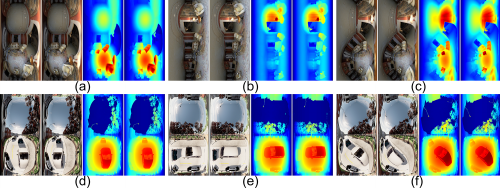}
  \caption{The qualitative results of the proposed omnidirectional stereo matching network on different camera rigs. (a)-(c) show the results of left-right, up-down and up-right pairs on 3D60. (d)-(f) show the results of 1-2, 1-3 and 1-4 pairs on Deep360. Each sample shows the left and right panoramas, the predicted disparity map and the ground truth,from left to right}\label{fig_disp_views}
\end{figure}

\subsection{Experiment Settings}\label{sec6_1}
\subsubsection{Datasets}\label{sec6_1_1}
We train and evaluate the networks on Deep360 and 3D60\cite{3D60} datasets to cover both indoor and outdoor scenes. The cameras rig of the Deep360 dataset consists of four 360$^\circ$ cameras set on a horizontal square. The 3D60 dataset employs a camera rig consisting of 360$^\circ$ cameras with up, center/left, and right views. We follow the official split of Deep360 dataset to evaluate the networks. We use one of the official dataset splits of 3D60\cite{3D60} that contains 7858 frames for training, 1103 for validation, and 2189 for testing in experiments. Furthermore, we evaluate the performance of our approaches on soiled data and compare the results across different numbers of views to demonstrate the adaptability and robustness of proposed methods.

Our experiments encompass the evaluation of the first stage of PSMODE for omnidirectional stereo matching and the evaluation of the full PSMODE and SSMODE for 360$^\circ$ depth estimation. For omnidirectional stereo matching, we present the results in the GEER projection, as the disparity is defined within the GEER domain. For a more comprehensive comparison of the depth estimation results with other methods, we display the depth results in the widely used ERP.

\subsubsection{Implementation Details}\label{sec6_1_2}
We implement both two-stage and one-stage frameworks with PyTorch. For the two-stage PSMODE network, we train the omnidirectional stereo matching network and depth fusion network independently. We first train the stereo matching network for 45 epochs with a learning rate of 0.001, and then decay the learning rate to 0.0001 to train the model for additional 10 epochs. For the depth fusion network of PSMODE, we train the network for 150 epochs with a learning rate of 0.0001. To evaluate the performance of PSMODE on soiled data, we further fine-tune the fusion network for 20 epochs on the soiled version of Deep360. For the SSMODE network, the initial training involved 45 epochs with a learning rate of 0.001, followed by 10 epochs with a learning rate of 0.0001. To evaluate the SSMODE network on the soiled version of the Deep360 dataset, we performed fine-tuning for 40 epochs with a learning rate of 0.00001. We set the depth range of SSMODE to $[0.5, 1000]$ meters and the number of hypothetical spheres to 192.

For the Deep360 dataset, we set the reference point of the depth map to the position of camera 1, while for the 3D60 dataset, we set the reference point of the depth map to the position of left/down camera. All SOTA 360$^\circ$ depth estimation methods are fine-tuned to achieve the best performance on each dataset. There is no result of OmniMVS on the 3D60 dataset due to the difference between the camera rigs.

\begin{figure}[t]%
  \centering
  \includegraphics[width=0.5\textwidth]{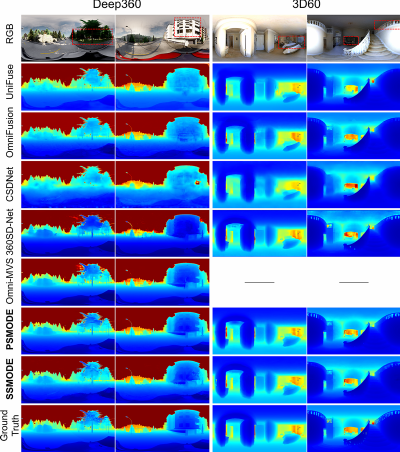}
  \caption{Qualitative results of PSMODE and SSMODE with other representative omnidirectional depth estimation methods on Deep360 (clear) and 3D60. We show the results on the widely used ERP projection. There is no result of OmniMVS on 3D60 due to the different input format}\label{fig_res_depth}
\end{figure}

\begin{figure}[h]%
  \centering
  \includegraphics[width=0.5\textwidth]{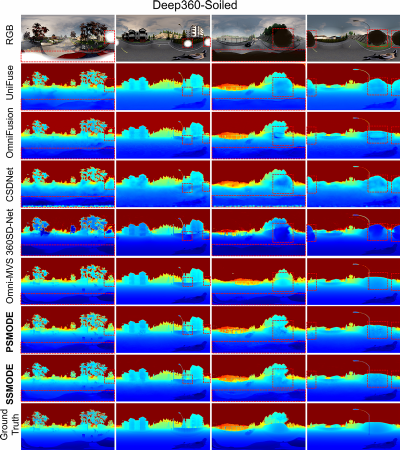}
  \caption{Qualitative results of PSMODE and SSMODE with other representative omnidirectional depth estimation methods on Deep360 (soiled). The proposed methods show higher robustness against camera soiling}\label{fig_res_soiled}
\end{figure}

\subsubsection{Metrics}\label{sec6_1_3}

We adopt two sets of metrics to evaluate the predicted disparity and depth results quantitatively. We use MAE (mean absolute error), RMSE (root mean square error), $Px1,3,5$(percentage of outliers with $pixel\ error > 1,3,5$), D1(percentage of outliers with $pixel\ error > 3\ and >5\%$)\cite{KITTI2015} to evaluate the disparity results. And we use MAE, RMSE, AbsRel (absolute relative error), SqRel (square relative error), SILog (scale-invariant logarithmic error)\cite{Silog}, $\delta 1,2,3$ (accuracy with threshold that $\max(\frac{\hat{y}}{y},\frac{y}{\hat{y}}) < 1.25,1.25^2,1.25^3$)\cite{Delta123} to evaluate the depth results. Higher values are better for the accuracies $\delta 1,2,3$, while lower values are better for other error metrics.

\subsection{Omnidirectional Stereo Matching}\label{sec6_2}
The existing binocular stereo matching algorithm is able to directly predict the spherical binocular disparity map at arbitrary relative positions based on the GEER projection method. Thus, we first evaluate the proposed omnidirectional stereo matching network on the Deep360 dataset and compare it with the excellent stereo matching algorithms PSMNet\cite{PSMNet}, AANet \cite{Xu2020aanet} and CREStereo\cite{Li2022crestereo}, as well as the omnidirectional method 360SD-Net\cite{360SD-Net}. For these approaches, we use the pre-trained models from the authors and follow their hyperparameters to finetune on Deep360. Fig. \ref{fig_res_disp} shows the qualitative results of omnidirectional stereo matching on the deep360 dataset. The quantitative results in Table \ref{tab_res_disp} illustrate that our stereo matching network with spherical feature learning achieves SOTA performance on 360$^\circ$ stereo matching.

We also present the results of stereo matching of two 360$^\circ$ cameras at different relative positions in Fig. \ref{fig_disp_views}. As shown in Fig. \ref{fig_res_disp} and Fig. \ref{fig_disp_views}, the proposed GEER projection establishes the epipolar constraint of binocular 360$^\circ$ cameras at arbitrary relative positions. The results show that the proposed stereo matching method with spherical feature extraction
module achieves high precision with clear details.

\subsection{Omnidirectional Depth Estimation}\label{sec6_3}
We evaluate the proposed PSMODE and SSMODE with SOTA omnidirectional depth estimation methods. To present the performance of SOTA works on Deep360, we test different types of methods, including monocular methods UniFuse\cite{Jiang2021unifuse} and omniFusion\cite{Li2022omnifusion}, binocular CSDNet\cite{Li2021csdnet} and 360SD-Net\cite{360SD-Net}, and multi-view OmniMVS\cite{Omnimvs_J}. All these models are fine-tuned with the pre-trained models from the authors. For the evaluation of the robustness of camera soiling, we finetune the models on the soiled version Deep360.

\begin{table*}[]
  \center
  \caption{Quantitative results of omnidirectional depth estimation on the proposed Deep360 dataset. The metrics refer to depth errors}
  \label{tab_res_depth}
  \resizebox{\textwidth}{!}{
    \begin{tabular}{cccccccccc}
      \toprule
      \multirow{2}{*}{Datasets}                           & \multirow{2}{*}{Methods}          & \multicolumn{8}{c}{Metrics}                                                                                                                                          \\ \cmidrule{3-10}
                                                          &                                   & MAE$\downarrow$             & RMSE$\downarrow$ & AbsRel$\downarrow$ & SqRel$\downarrow$ & SILog$\downarrow$ & Delta1$\uparrow$ & Delta2$\uparrow$ & Delta3$\uparrow$ \\ \midrule
      \multirow{8}{*}{Deep360}                            & Unifuse\cite{Jiang2021unifuse}    & 3.9193                      & 28.8475          & 0.0546             & 0.3125            & 0.1508            & 96.0269          & 98.2679          & 98.9909          \\
                                                          & OmniFusion\cite{Li2022omnifusion} & 7.6873                      & 45.8307          & 0.1374             & 2.5297            & 0.2348            & 94.5733          & 97.8327          & 98.5763          \\
                                                          & CSDNet\cite{Li2021csdnet}         & 6.6548                      & 36.5526          & 0.1553             & 1.7898            & 0.2475            & 86.0836          & 95.1589          & 97.7562          \\
                                                          & 360SD-Net\cite{360SD-Net}         & 11.2643                     & 66.5789          & 0.0609             & 0.5973            & 0.2438            & 94.8594          & 97.2050          & 98.1038          \\
                                                          & OmniMVS\cite{Omnimvs_J}           & 8.8865                      & 59.3043          & 0.1073             & 2.9071            & 0.2434            & 94.9611          & 97.5495          & 98.2851          \\
                                                          & PSMODE(w/o fusion)                & 7.7024                      & 52.1627          & 0.0412             & 0.5244            & 0.1944            & 96.8257          & 98.1596          & 98.7035          \\
                                                          & PSMODE                            & \textbf{3.2483}             & \textbf{24.9391} & \textbf{0.0365}    & \textbf{0.0789}   & \textbf{0.1104}   & \textbf{97.9636} & \textbf{99.0987} & \textbf{99.4683} \\
                                                          & SSMODE                            & 4.7118                      & 38.6426          & 0.0590             & 0.5318            & 0.2099            & 95.1759          & 97.9139          & 98.6693          \\ \midrule
      \multicolumn{1}{l}{\multirow{8}{*}{Deep360-Soiled}} & Unifuse\cite{Jiang2021unifuse}    & 5.4636                      & 37.4313          & 0.1119             & 4.8948            & 0.1810            & 95.2379          & 97.8686          & 98.7208          \\
      \multicolumn{1}{l}{}                                & OmniFusion\cite{Li2022omnifusion} & 8.5136                      & 49.3830          & 0.1471             & 3.0937            & 0.2471            & 93.8283          & 97.5569          & 98.4261          \\
      \multicolumn{1}{l}{}                                & CSDNet\cite{Li2021csdnet}         & 7.5950                      & 38.4693          & 0.1631             & 3.7148            & 0.2521            & 86.7329          & 95.3295          & 97.7513          \\
      \multicolumn{1}{l}{}                                & 360SD-Net\cite{360SD-Net}         & 22.5495                     & 97.3958          & 0.1060             & 1.1857            & 0.4465            & 90.5868          & 94.1468          & 98.6262          \\
      \multicolumn{1}{l}{}                                & OmniMVS\cite{Omnimvs_J}           & 9.2680                      & 62.1838          & 0.1935             & 22.6994           & 0.2597            & 94.7009          & 97.3821          & 98.1652          \\
      \multicolumn{1}{l}{}                                & PSMODE(w/o fusion)                & 15.2145                     & 77.5905          & 0.1230             & 6.3135            & 0.5466            & 93.2377          & 96.0349          & 97.1837          \\
      \multicolumn{1}{l}{}                                & PSMODE                            & \textbf{4.4652}             & \textbf{31.7124} & \textbf{0.0495}    & \textbf{0.1778}   & \textbf{0.1458}   & \textbf{96.3504} & \textbf{98.5718} & \textbf{99.2109} \\
      \multicolumn{1}{l}{}                                & SSMODE                            & 5.0007                      & 40.2564          & 0.0667             & 0.7543            & 0.2179            & 94.4836          & 97.7393          & 98.6033          \\ \bottomrule
    \end{tabular}%
  }
\end{table*}

As shown in Table \ref{tab_res_depth}, PSMODE and SSMODE perform favorably against SOTA omnidirectional depth estimation methods, especially on the dataset with soiled panoramas. We also compare the result of PSMODE with and without the fusion stage in Table \ref{tab_res_depth}. As the results show, the multi-view depth fusion stage significantly improves the accuracy of omnidirectional depth estimation.
As demonstrated in Table \ref{tab_res_depth} and Fig. \ref{fig_res_soiled}, the accuracy degradation of the proposed methods on the soiled data is significantly lower than that of existing methods. The comparison demonstrates the robustness of the proposed multi-view depth estimation methods against camera soiling.
We also evaluate the proposed methods on 3D60 dataset and illustrate the results in Table \ref{tab_res_3d60} and Fig. \ref{fig_res_depth}. The proposed PSMODE and SSMODE achieve high accuracy on both indoor and outdoor scenes. In this paper, we leverage all three stereo pairs within the 3D60 (left-right,up-down,up-right) in the depth fusion stage of PSMODE by employing the GEER projection. Thus, the results in Table \ref{tab_res_3d60} is better than those reported in the conference version\cite{Li2022mode}.

\begin{table*}[]
  \center
  \setlength{\tabcolsep}{0.7mm}
  \caption{Quantitative results of omnidirectional depth estimation on 3D60 dataset. The metrics refer to depth errors}
  \label{tab_res_3d60}
  \resizebox{0.7\textwidth}{!}{%
    \begin{tabular}{ccccccccc}
      \toprule
      \multirow{2}{*}{Methods}          & \multicolumn{8}{c}{Metrics}                                                                                                                                          \\ \cmidrule{2-9}
                                        & MAE$\downarrow$             & RMSE$\downarrow$ & AbsRel$\downarrow$ & SqRel$\downarrow$ & SILog$\downarrow$ & Delta1$\uparrow$ & Delta2$\uparrow$ & Delta3$\uparrow$ \\ \midrule
      Unifuse\cite{Jiang2021unifuse}    & 0.1868                      & 0.3947           & 0.0799             & 0.0246            & 0.1126            & 93.2860          & 98.4839          & 99.4828          \\
      omniFusion\cite{Li2022omnifusion} & 0.1521                      & 0.3297           & 0.0628             & 0.0138            & 0.0892            & 96.0063          & 99.2099          & 99.7610          \\
      CSDNet\cite{Li2021csdnet}         & 0.2067                      & 0.4225           & 0.0908             & 0.0241            & 0.1273            & 91.9537          & 98.3936          & 99.5109          \\
      360SD-Net\cite{360SD-Net}         & 0.0762                      & 0.2639           & 0.0300             & 0.0117            & 1.4578            & 97.6751          & 98.6603          & 99.0417          \\
      PSMODE                            & \textbf{0.0619}             & \textbf{0.1837}  & \textbf{0.0236}    & \textbf{0.0033}   & \textbf{0.0426}   & \textbf{99.3806} & \textbf{99.8584} & \textbf{99.9452} \\
      SSMODE                            & 0.0753                      & 0.2422           & 0.0300             & 0.0098            & 0.0638            & 98.4621          & 99.5247          & 99.8002          \\ \bottomrule
    \end{tabular}%
  }
\end{table*}

\subsection{Results on Real Scenes}\label{sec6_4}
\begin{figure}[t]%
  \centering
  \includegraphics[width=0.45\textwidth]{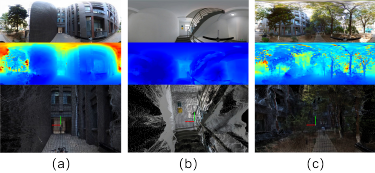}
  \caption{Predict depth maps and point clouds on real-scene data. Each row from top to bottom represents the panorama, the predicted depth, and the point cloud of the front view, respectively. We use the best model of PSMODE trained on Deep360 for real-scene inference}\label{fig_res_real}
\end{figure}

We use the best PSMODE model trained on Deep360 to predict 360$^\circ$ depth maps on real-scene data. We use four Insta One X2 360$^\circ$ cameras to build the camera system, as shown in Fig. \ref{fig_overview}(b). Fig. \ref{fig_res_real} illustrates that the proposed algorithm also achieves an accurate depth estimation on real-scene data.

\subsection{Evaluation of Different Numbers of Views}\label{sec6_5}
The proposed PSMODE fuses depth maps estimated from various stereo pairs, While SSMODE constructs a spherical cost volume based on panorama features. Both the PSMODE and SSMODE frameworks are designed to accommodate different numbers of views, offering flexibility in terms of the input camera configurations.

\begin{figure}[t]%
  \centering
  \includegraphics[width=0.5\textwidth]{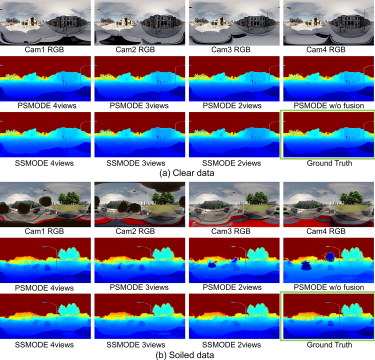}
  \caption{Comparison of PSMODE and SSMODE with different view numbers. (a) and (b) show the results on clear data and soiled data, respectively}\label{fig_diff_views}
\end{figure}

To evaluate the performance of PSMODE and SSMODE under varying view conditions, we conducted experiments on the clear and soiled Deep360 dataset using different numbers of views (4, 3, 2). Table \ref{tab_diff_views} and Fig. \ref{fig_diff_views} indicate that as the number of views decreases, both PSMODE and SSMODE experience increase in error of depth estimation. While PSMODE achieves higher accuracy on normal data, its accuracy decline on soiled data is more pronounced when using fewer cameras. In contrast, SSMODE demonstrates greater robustness against soiled data with a reduced number of views. The qualitative results in Fig. \ref{fig_diff_views} illustrate that PSMODE predicts more detailed and accurate depth information, while SSMODE exhibits better performance on soiled data with only 2 views.

Moreover, compared with the results of existing methods in Table \ref{tab_res_depth}, PSMODE and SSMODE achieve comparable performance with only 2 views. The experiments demonstrate that the proposed two frameworks are compatible with different numbers of views.

\subsection{Ablation Study}\label{sec6_6}
We leverage spherical convolution in the feature extraction module and remove the image cropping during training PSMODE. We also add RGB panoramas and confidence maps into the depth fusion network. To verify the improvement of each component, we adopt ablation experiments on the two stages of SSMODE. Table \ref{tab_ablation_disp} shows the ablation studies of the omnidirectional stereo matching network. The results show that using panoramas without cropping and applying spherical convolution improve the performance. Table \ref{tab_ablation_depth} illustrates the ablation studies of the depth map fusion network. The results show that the fusion stage improves the quality of depth maps. The rows of the table gradually show the improvement of adding each component into the network.

\begin{table*}[]
  \center
  \caption{Quantitative results of PSMODE and SSMODE with different view numbers on Deep360 dataset. The metrics refer to depth errors}
  \label{tab_diff_views}
  \resizebox{0.9\textwidth}{!}{%
    \begin{tabular}{ccccccccccc}
      \toprule
      \multirow{2}{*}{Datasets}                           & \multirow{2}{*}{Methods} & \multirow{2}{*}{Num of Views} & \multicolumn{8}{c}{Metrics}                                                                    \\ \cmidrule{4-11}
                                                          &
                                                          &
                                                          &
      \multicolumn{1}{c}{MAE$\downarrow$}                 &
      \multicolumn{1}{c}{RMSE$\downarrow$}                &
      \multicolumn{1}{c}{AbsRel$\downarrow$}              &
      \multicolumn{1}{c}{SqRel$\downarrow$}               &
      \multicolumn{1}{c}{SILog$\downarrow$}               &
      \multicolumn{1}{c}{Delta1$\uparrow$}                &
      \multicolumn{1}{c}{Delta2$\uparrow$}                &
      \multicolumn{1}{c}{Delta3$\uparrow$}                                                                                                                                                                            \\ \midrule
      \multirow{6}{*}{Deep360}                            & \multirow{3}{*}{PSMODE}  & 4                             & 3.2483                      & 24.9391 & 0.0365 & 0.0789 & 0.1104 & 97.9636 & 99.0987 & 99.4683 \\
                                                          &                          & 3                             & 3.8269                      & 32.1204 & 0.0456 & 0.3243 & 0.1473 & 97.5363 & 98.8140 & 99.2348 \\
                                                          &                          & 2                             & 3.9357                      & 33.1037 & 0.0533 & 0.3953 & 0.1568 & 97.1295 & 98.7424 & 99.1972 \\ \cmidrule{2-11}
                                                          & \multirow{3}{*}{SSMODE}  & 4                             & 4.7118                      & 38.6426 & 0.0590 & 0.5318 & 0.2099 & 95.1759 & 97.9139 & 98.6693 \\
                                                          &                          & 3                             & 4.7579                      & 38.7975 & 0.0608 & 0.5349 & 0.2114 & 95.0128 & 97.8555 & 98.6394 \\
                                                          &                          & 2                             & 4.7726                      & 38.8260 & 0.0619 & 0.5436 & 0.2135 & 94.9300 & 97.8580 & 98.6390 \\ \midrule
      \multicolumn{1}{c}{\multirow{6}{*}{Deep360-Soiled}} &
      \multirow{3}{*}{PSMODE}                             &
      4                                                   &
      4.4652                                              &
      31.7124                                             &
      0.0495                                              &
      0.1778                                              &
      0.1458                                              &
      96.3504                                             &
      98.5718                                             &
      99.2109                                                                                                                                                                                                         \\
      \multicolumn{1}{c}{}                                &                          & 3                             & 5.6072                      & 39.6076 & 0.0795 & 0.6459 & 0.1846 & 94.6837 & 97.8830 & 98.7619 \\
      \multicolumn{1}{c}{}                                &                          & 2                             & 5.9115                      & 41.8285 & 0.0819 & 1.4762 & 0.2054 & 94.5810 & 97.5135 & 98.4756 \\ \cmidrule{2-11}
      \multicolumn{1}{c}{}                                & \multirow{3}{*}{SSMODE}  & 4                             & 5.0007                      & 40.2564 & 0.0667 & 0.7543 & 0.2179 & 94.4836 & 97.7393 & 98.6033 \\
      \multicolumn{1}{c}{}                                &                          & 3                             & 5.1032                      & 40.5233 & 0.0697 & 0.6361 & 0.2223 & 93.8133 & 97.5211 & 98.5026 \\
      \multicolumn{1}{c}{}                                &                          & 2                             & 5.2049                      & 41.1470 & 0.0770 & 1.3269 & 0.2267 & 93.3870 & 97.4780 & 98.5021 \\ \bottomrule
    \end{tabular}%
  }
\end{table*}

\begin{table*}[]
  \centering
  \caption{Ablation studies for omnidirectional stereo matching on Deep360. We compare the results of the proposed network with and without Input Image Cropping (\textbf{Cr}) and Spherical Convolution (\textbf{SC}). The metrics refer to disparity errors}
  \label{tab_ablation_disp}
  \setlength{\tabcolsep}{1.0mm}
  \resizebox{0.5\textwidth}{!}{%
    \begin{tabular}{cc|cccccc}
      \toprule
      \multicolumn{2}{c|}{\begin{tabular}[c]{@{}c@{}}Network\\ settings\end{tabular}} & \multicolumn{6}{c}{Metrics}                                                                                                                             \\ \midrule
      \textbf{Cr}                                                                     & \textbf{SC}                 & MAE$\downarrow$ & RMSE$\downarrow$ & Px1(\%)$\downarrow$ & Px3(\%)$\downarrow$ & Px5(\%)$\downarrow$ & D1(\%)$\downarrow$ \\ \midrule
      \checkmark                                                                      & $\times$                    & 0.3220          & 1.7425           & 3.9787              & 1.3042              & 0.8049              & 1.2588             \\
      $\times$                                                                        & $\times$                    & 0.2109          & 1.2408           & 2.6509              & 0.8967              & 0.5377              & 0.8846             \\
      $\times$                                                                        & \checkmark                  & \textbf{0.2073} & \textbf{1.2347}  & \textbf{2.6010}     & \textbf{0.8767}     & \textbf{0.5260}     & \textbf{0.8652}    \\ \bottomrule
    \end{tabular}
  }
\end{table*}

\begin{table*}[]
  \center
  \setlength{\tabcolsep}{0.55mm}
  \caption{Ablation studies for multi-view depth fusion in PSMODE on soiled Deep360. We compare the performance of the proposed fusion network with and without RGB images and Confidence maps. We list the result that without fusion (w.r.t the results of stereo matching stage in PSMODE) in the first row as the baseline. The metrics refer to depth errors}
  \label{tab_ablation_depth}
  \resizebox{0.7\textwidth}{!}{%
    \begin{tabular}{ccc|cccccccc}
      \toprule
      \multicolumn{3}{c|}{Network settings}  & \multicolumn{8}{c}{Metrics}                                                                                                                                                                                        \\ \midrule
      \multicolumn{1}{c}{fusion}             &
      Img                                    &
      \multicolumn{1}{c|}{Conf}              &
      \multicolumn{1}{c}{MAE$\downarrow$}    &
      \multicolumn{1}{c}{RMSE$\downarrow$}   &
      \multicolumn{1}{c}{AbsRel$\downarrow$} &
      \multicolumn{1}{c}{SqRel$\downarrow$}  &
      \multicolumn{1}{c}{SILog$\downarrow$}  &
      \multicolumn{1}{c}{Delta1$\uparrow$}   &
      \multicolumn{1}{c}{Delta2$\uparrow$}   &
      \multicolumn{1}{c}{Delta3$\uparrow$}                                                                                                                                                                                                                        \\ \midrule
      \multicolumn{1}{c}{$\times$}           & \multicolumn{1}{c}{$\times$} & $\times$                        & 15.2145         & 77.5905          & 0.1230          & 6.3135          & 0.5466          & 93.2377          & 96.0349          & 97.1837          \\
      \checkmark                             & \multicolumn{1}{c}{$\times$} & $\times$                        & 6.2548          & 45.8603          & 0.0516          & 0.2702          & 0.1831          & 95.9953          & 98.1431          & 98.8211          \\
      \checkmark                             & \checkmark                   & $\times$                        & \textbf{4.2071} & 32.0112          & 0.0710          & 0.2443          & 0.1554          & 95.1875          & 98.4766          & 99.1773          \\
      \checkmark                             & \checkmark                   & \multicolumn{1}{c|}{\checkmark} & 4.4652          & \textbf{31.7124} & \textbf{0.0495} & \textbf{0.1778} & \textbf{0.1458} & \textbf{96.3504} & \textbf{98.5718} & \textbf{99.2109} \\ \bottomrule
    \end{tabular}
  }
\end{table*}

\subsection{Comparison of Two-stage and One-stage Methods}\label{sec6_7}
As illustrated in Table \ref{tab_res_depth} and Table \ref{tab_res_3d60}, PSMODE outperforms SSMODE on the Deep360 dataset when utilizing four cameras. According to the results in Table \ref{tab_diff_views} the accuracy of PSMODE experiences a more significant decrease on soiled data when the number of cameras decreases. PSMODE fuses the results of pairwise stereo matching, which can integrate the information of different views to mitigate the distortion and blind points of the GEER projection. Consequently, the number of views has a more pronounced impact on PSMODE. On the other hand, SSMODE constructs a unified cost volume for all cameras and exhibits slightly lower accuracy compared to PSMODE. However, SSMODE demonstrates greater robustness to variations in the number of input cameras.

We also compare the video memory usage and time consumption of PSMODE and SSMODE, with the details provided in Table \ref{tab_comp2methods}. The two-stage PSMODE consists of an omnidirectional stereo matching network and a depth fusion network, and both networks can be trained independently. Therefore, PSMODE can employ a larger model with more video memory. However, the two-stage pipeline of PSMODE costs more time during the inference phase. SSMODE requires more video memory in training but has a faster inference speed. Moreover, PSMODE needs to estimate the depth map for each camera pair by stereo matching, which increases the computational complexity. As the number of cameras increases, the computational complexity of PSMODE grows significantly, resulting in reduced efficiency of the method.

In summary, the two-stage PSMODE achieves higher accuracy performance, and can also achieve larger parameters by training two networks independently. The one-stage SSMODE is more robust to changes in the number of cameras and more efficient at the inference phase, especially when the number of cameras is large.
\begin{table}[]
  \caption{Comparison of PSMODE and SSMODE in training memory and inference time. We use NVIDIA RTX3090 for training and inference, and set the resolution of input panoramas as 1024$\times$512 and batch size as one}\label{tab_comp2methods}
  \begin{tabular}{ccc}
    \toprule
    Methods & Training video mem.                                                                  & Inference time. \\ \midrule
    PSMODE  & \begin{tabular}[c]{@{}c@{}}13GB (stereo matching)\\  4GB (depth fusion)\end{tabular} & 1.85 s/frame    \\ \midrule
    SSMODE  & 19GB                                                                                 & 0.32 s/frame    \\ \bottomrule
  \end{tabular}%
\end{table}

\section{Discussion and Conclusion}\label{sec7}
\subsection{GEER projection}\label{sec7_1}
As shown in Fig. \ref{fig_projection}(a), we transform the panoramas into GEER projection to build the epipolar constraint for binocular 360$^\circ$ cameras and represent the disparity with the angle difference. However, for those points on the x-axis (line $O_l O_r$), the angle $\phi$ is always the same on left and right cameras:
\begin{align}
  \phi_l^p=\phi_r^p = {0} \ or \ {\pi}, p \in ([x,0,0],-\infty < x < \infty)\label{eq8}
\end{align}
Thus, there is no angle difference or disparity for the points on the x-axis. These points are located in the leftmost column and the rightmost column of the GEER projection images, which we call blind points. In summary, the GEER projection establishes the epipolar constraint for binocular panorama pairs, but it is difficult to estimate the accuracy depth value of the blind points.

\subsection{Conclusion}\label{sec7_3}
In this paper, we focus on the multi-view omnidirectional depth estimation(MODE) with multiple 360$^\circ$ cameras. We leverage the geometry constraint and redundant information of multi-view panoramas to enhance robustness against camera soiling caused by factors such as mud, water drops, or intense glare. We propose the two-stage PSMODE approach based on pairwise stereo matching and fusion, and the one-stage SSMODE approach based on spherical sweeping. Experiments demonstrate that both two approaches achieve SOTA performance and can predict high quality depth maps with soiled panoramas. We also validate the flexibility and compatibility of the rigs and numbers of cameras for both two methods.

In practical applications, fisheye cameras are often more prevalent than 360$^\circ$ cameras\cite{Qian2022FisheyeSurvey}. We consider fisheye images as partially occluded spherical images. Thus, the proposed Generalized Epipolar Equirectangular (GEER) projection and depth estimation algorithms are applicable to this setting. However, fisheye cameras have smaller field-of-view (FoV) and exhibit limited overlapping areas between cameras when compared to 360$^\circ$ cameras. Notably, PSMODE requires a larger overlapping area since it relies on stereo matching to obtain initial depth maps. SSMODE also requires a common field of view for the cameras, and areas where only one camera is visible will lead to degraded monocular depth estimation. Consequently, the processing of overlapping and non-overlapping areas emerges as an open problem in multi-view omnidirectional depth estimation. Furthermore, we will study the real-time optimization of the algorithms in future work to improve the efficiency of 3D measurement in practical.

\bibliography{main}

\end{document}